\documentclass{article}

 \usepackage[main,final]{iaseai26}

\usepackage[utf8]{inputenc}
\usepackage[T1]{fontenc}
\usepackage{hyperref}
\usepackage{url}
\usepackage{booktabs}
\usepackage{amsfonts}
\usepackage{amsmath}
\usepackage{amssymb}
\usepackage{nicefrac}
\usepackage{microtype}
\usepackage{xcolor}
\usepackage{mathtools, amsthm}
\usepackage{bm}
\usepackage{cleveref}
\usepackage{tikz}
\usetikzlibrary{arrows.meta}
\usetikzlibrary{positioning, shapes.multipart, shapes.misc, arrows, backgrounds,calc,patterns}
\usetikzlibrary{positioning,shapes,arrows,calc,decorations.pathreplacing}
\usepackage{comment}
\usepackage{bbm}
\usepackage{wrapfig}
\usepackage{caption}
\usepackage{subcaption}
\usepackage{changepage}
\usepackage{graphicx}
\usepackage[table]{xcolor}
\usepackage{multirow}
\usepackage{algorithm}
\usepackage{algpseudocode}
\usepackage[many]{tcolorbox}
\usepackage{natbib}
\setcitestyle{aysep={}}
\usepackage[toc,page]{appendix}
\usepackage{listings}

\DeclareMathOperator*{\argmax}{arg\,max}

\crefname{section}{§\hspace{-0.1cm}}{§§}
\Crefname{section}{§}{§§}

\tcbset{
    sharp corners,
    colback = white,
    before skip = 0.2cm,
    after skip = 0.5cm,
    breakable,
    height fixed for=none
}

\newtcolorbox{boxA}{
    boxrule = 1.5pt,
    colframe = black,
    rounded corners,
    arc = 5pt,
    breakable
}

  {\list{}{\leftmargin=#1\rightmargin=#1}\item[]}%
  {\endlist}

\def\Frame{\langle \Omega, \mathcal{F}, \pi, \mathcal{P}\rangle}
\def\A{\mathcal{A}}

\def\E{\mathbb{E}}
\def\P{\mathcal{P}}
\def\F{\mathcal{F}}

\theoremstyle{plain}
\newtheorem{theorem}{Theorem}[section]

\theoremstyle{definition}
\newtheorem{defn}[theorem]{Definition}

\theoremstyle{remark}

\newcounter{daggerfootnote}

\makeatletter
\newcommand{\equalcontrib}{\textsuperscript{\dag}}
\makeatother

\title{A Decision-Theoretic Approach for Managing Misalignment}

\author{%
  Daniel A. Herrmann\equalcontrib\\
  University of North Carolina at Chapel Hill\\
  \texttt{danher@unc.edu}
  \And
  Abinav Chari\\
  Georgia Institute of Technology\\
  \texttt{abinav.ram.chari@gmail.com}
  \And
  Isabelle Qian\\
  University of California, Berkeley\\
  \texttt{isabelle.qian@gmail.com}
   \And
 Sree Sharvesh\\
  Amrita Vishwa Vidyapeetham\\
  \texttt{sssreesharvesh@gmail.com}
  \And
  B. A. Levinstein\equalcontrib\\
  Anthropic\\
  University of Illinois at Urbana-Champaign\\
  \texttt{benlevin@illinois.edu}
}

\begin{document}

\maketitle

\begingroup
  \renewcommand\thefootnote{\dag}
   \footnotetext{Equal contribution.}
\endgroup

\begin{abstract}
When should we delegate decisions to AI systems? While the value alignment literature has developed techniques for shaping AI values, less attention has been paid to how to determine, under uncertainty, when imperfect alignment is good enough to justify delegation. We argue that rational delegation requires balancing an agent's value (mis)alignment with its epistemic accuracy and its reach (the acts it has available). This paper introduces a formal, decision-theoretic framework to analyze this tradeoff precisely accounting for a principal's uncertainty about these factors. Our analysis reveals a sharp distinction between two delegation scenarios. First, universal delegation (trusting an agent with any problem) demands near-perfect value alignment and total epistemic trust, conditions rarely met in practice. Second, we show that context-specific delegation can be optimal even with significant misalignment. An agent's superior accuracy or expanded reach may grant access to better overall decision problems, making delegation rational in expectation. We develop a novel scoring framework to quantify this ex ante decision. Ultimately, our work provides a principled method for determining when an AI is aligned enough for a given context, shifting the focus from achieving perfect alignment to managing the risks and rewards of delegation under uncertainty.
\end{abstract}

\section{Introduction}

When should we delegate decisions to AI systems? This is one of the key questions in AI safety, as we must decide when we expect things to go better for ourselves and humanity by outsourcing decision-making to AI. 
An intuitive answer implicit in much of the value alignment literature is: when we've successfully aligned their values with our own. But this answer overlooks two critical facts about real-world delegation.

First, we routinely delegate to agents whose values differ from ours. Many people delegate investment decisions to a financial advisor who charges fees they'd rather not pay. We delegate medical decisions to doctors who spend less time thinking about our specific case than we'd wish them to. These delegations make sense because the agents' superior competence outweighs their value misalignment.

Second, we make these delegation decisions under uncertainty. We cannot be certain about either an agent's values or their capability. Any practical framework for AI delegation must account for this uncertainty and help us manage it.

These observations reveal a gap in the value alignment literature. Significant effort has gone into developing alignment techniques such as reinforcement learning with human feedback and constitutional AI. These techniques are aimed at answering the question: how do we nudge AI values in the direction of our own values? Less attention has been paid to the complementary question: how do we determine when an AI system is \textit{sufficiently} aligned to warrant delegation? This is not simply a matter of measuring how well we shaped the values of a system. It requires weighing the misalignment against the capability under uncertainty.

In this article we provide a formal framework that makes this tradeoff precise by modeling three key factors: the agent's \textbf{epistemic accuracy} (how well-calibrated its beliefs are), its \textbf{value alignment} (how closely its objectives match ours), and its \textbf{reach} (what kinds of problems it can access and solve). Our framework treats delegation as a decision problem where a principal must choose whether to delegate based on her beliefs about all three factors.

Our analysis yields a sharp contrast between two types of delegation. \textbf{Universal delegation}---being willing to delegate any decision problem to the agent---requires an extremely demanding condition: the principal must totally trust the agent's beliefs when values are aligned, and near-perfect alignment when values differ. This helps explain why full automation might be inadvisable even for capable AI systems. However, \textbf{context-specific delegation}---delegating particular types of problem or accepting delegation in expectation---can be rational even with moderate misalignment, provided the agent's superior epistemic accuracy or expanded reach generates sufficient gains.

We consider three important results. First, as shown in \cite{dorst2021deference}, when the principal and agent share values and face the same problems, universal delegation is equivalent to a condition we call total trust: the principal must defer completely to the agent's expectations (Theorem \ref{thm:dorst}). Second, we prove that when values differ but problem distributions remain shared, universal delegation requires near-perfect alignment (Theorem \ref{thm:benfav}). Third, we show that when the agent's greater reach changes the distribution of problems it faces, delegation becomes an expected value calculation that can favor a misaligned but capable agent over perfect alignment with limited reach.

We build our framework in three stages that isolate each factor's contribution. First, we analyze delegation when the principal and agent share values and face the same problems, differing only in beliefs. This establishes the baseline epistemic requirements for delegation. Next, we allow for value misalignment while maintaining shared problem distributions. This reveals how misalignment constrains delegation even with superior capability. Finally, we incorporate how an agent's greater reach changes the distribution of problems they face. This shows when expanded capabilities can overcome value misalignment. We conclude by discussing implications for AI deployment decisions and identifying key empirical questions for applying the framework.

Throughout this paper, we assume both principal and agent are Bayesian expected utility maximizers. While this assumption rarely holds in practice, particularly for neural network-based AI systems that may exhibit non-transitive preferences or context-dependent behavior, our analysis remains informative for two reasons. First, it provides a normative benchmark for rational delegation. Even though humans deviate from Bayesian ideals, understanding optimal delegation can guide our decisions about when to trust AI systems with important tasks. We can use these theoretical insights as anchors while adjusting for our own cognitive limitations and the specific contexts we face. Second, our results establish lower bounds on the alignment challenge. When we show that universal delegation requires near-perfect alignment even with transparent, interpretable Bayesian agents (\ref{thm:benfav}), we're identifying fundamental tensions that persist regardless of architectural details. If achieving sufficient alignment for broad delegation is difficult when both parties' decision processes are well-understood and rational, the challenge only intensifies with opaque systems whose objectives and internal representations remain unclear. Thus, while future work should extend this framework to non-Bayesian agents, the current analysis already reveals why the delegation problem is so thorny.

\section{Related Works}

\textbf{Foundations of Rational Deference}. Our framework builds on established equivalences in the theory of rational deference. When agents \textit{reflect} experts, meaning they adopt their credences upon learning them, this epistemic deference proves equivalent to decision-theoretic deference under standard conditions. That is, an agent reflects an expert if and only if she willingly delegates decisions to that expert (\cite{skyrms1990value}; \cite{huttegger2014learning}). This delegation, in turn, occurs precisely when the agent expects the expert to be more accurate according to proper scoring rules (\cite{levinstein2017pragmatist}; \cite{campbell2024accuracy}). \cite{dorst2021deference} extends these results to weaker deference principles, establishing the \textbf{total trust} condition we discuss in \ref{del}. \cite{levinstein2023accuracy} and \cite{schervish1989general} formalize the accuracy-based evaluation of credences under different measure over decision problems. However, these frameworks assume \textit{aligned} values. Our key contribution is showing how value misalignment breaks these clean equivalences, requiring us to balance epistemic superiority against divergent objectives.

\textbf{Principal-Agent Theory and AI Delegation}. Classic principal-agent models provide insights about delegation under known parameters. \cite{aghion1997} shows delegation becomes attractive with informed, aligned agents, while \cite{dessein2002} demonstrates how alignment level determines optimal organizational structure. Yet these models assume the principal knows the agent's bias which is unrealistic for AI systems. Our framework addresses this by treating both alignment and capability as uncertain, analyzing when delegation remains rational despite this compound uncertainty.

\textbf{Scoring Frameworks for Decision Making}. Our technical approach deploys and extends \cite{konek2023evaluating}'s scoring framework. While Konek evaluates imprecise forecasts under a fixed distribution of problems, we allow distributions to vary between principal and agent. This captures a crucial insight: an agent's expanded reach changes not just decision quality but the decision landscape itself. The underlying framework of subjective expected utility (see  \cite{fishburn1981subjective} and \cite{schervish1991shared} for references) regiments our formal treatment.

\textbf{Bridging Theory and Practice}. Our framework provides theoretical foundations for empirical AI safety research. \cite{bhatt2025ctrl} empirically measure capability-safety trade-offs but lacks a general theory. Our scoring framework formalizes such trade-offs precisely. \cite{hadfield2016cooperative} frame alignment as cooperative, assuming shared (but initially unknown) objectives. We complement this by analyzing the prior question: given uncertainty about achieved alignment, when should delegation occur? For practical implementation where agent utilities must be inferred, inverse reinforcement learning provides essential machinery \cite{ng2000algorithms, abbeel2004apprenticeship, ramachandran2007bayesian, ziebart2008maximum}, though our results show that even with rich information about utilities, delegation requires careful analysis.

\textbf{The Alignment Sufficiency Problem}. While substantial effort targets improving AI alignment, less attention addresses determining when alignment \textit{suffices} for delegation. Our work fills this gap by providing principled methods to evaluate whether an AI system, given our imperfect information about its particular mix of alignment, accuracy, and reach, merits delegation in specific contexts. This shifts focus from the alignment process to the delegation decision, addressing a critical need as AI capabilities expand. 

\section{The Framework}
\label{framework}

We construct the delegation framework in three layers, each relaxing an idealization of the previous one. Section \ref{beliefs} isolates epistemic uncertainty (shared values, shared reach). Section \ref{value} introduces value uncertainty. Section \ref{reach} generalizes further to reach uncertainty, where the agent’s capabilities alter the problem distribution itself. This layered structure clarifies which assumptions drive each result.

\subsection{Uncertainty About the Agent's Beliefs}
\label{beliefs}

Here we hold values and reach fixed and isolate the epistemic dimension: when should a principal delegate purely on the basis of superior epistemic accuracy?

We model the delegation decision from the perspective of a Bayesian principal (the human) who must decide whether to delegate to an agent (the AI). The principal maximizes expected utility under uncertainty about both the state of the world and the agent's characteristics---particularly its beliefs and values.
The key modeling challenge is that the principal is uncertain about what the agent believes. To capture this, we use a probability frame that represents both the principal's beliefs and her uncertainty about the agent's beliefs:

\begin{defn}[Probability Frame]
    A \textbf{probability frame} is a quadruple $\Frame$, where:
    \begin{enumerate}
        \item $\langle \Omega, \F, \pi \rangle$ is a finite probability space representing the principal's beliefs, and
        \item $\P = \{P_{\omega}\}_{\omega\in\Omega}$ assigns to each state $\omega$ a probability function $P_{\omega}$ on $\F$ representing agent's beliefs at that state.  
    \end{enumerate}
\end{defn}

Intuitively, $\pi$ captures what the principal believes, while $P_{\omega}$ captures what the agent believes at state $\omega$. Since the principal is uncertain about the state, she is also uncertain about the agent's beliefs. 

We write $P$ (without subscript) to denote the agent's actual probability function, treating it as a state-dependent variable. Similarly, $\mathbb{E}$ denotes the agent's expectation operator. This allows us to express the principal's beliefs about the agent's beliefs directly. For instance, $[P(X)=p]$ denotes the set of states where the agent assigns probability $p$ to event $X$, and $\pi([P(X)=p])$ is the principal's probability that the agent assigns probability $p$ to $X$.

\subsubsection{Delegation}
\label{del}

Given our set-up of the principal uncertain about the agent's beliefs, we can ask: when should the principal delegate to the agent? To answer this, we first need to specify what decisions they face.

A \textbf{decision problem} is a set $\mathcal{O}$ of options, where each option $O:\Omega \to \mathbb{R}$ is a random variable representing payoffs. If the principal chooses for herself, she selects an option maximizing her expected payoff: some
$O^* \in \operatorname*{arg\,max}_{O \in \mathcal{O}} \mathbb{E}_\pi[O].$

If she delegates to the agent (who we assume shares her payoffs), the agent will choose the option maximizing \emph{its} expected payoff. We represent this with an \textbf{expert strategy} $S = \{S_{\omega}\}_{\omega\in\Omega}$ where
$S_\omega \in \operatorname*{arg\,max}_{O \in \mathcal{O}} \mathbb{E}_\omega[O]$. That is, at each state $\omega$, the agent chooses optimally according to beliefs $P_\omega$.

Consider first the strongest possible condition for delegation. We say the principal \textbf{values} the agent if delegation is always weakly preferred: for any decision problem $\mathcal{O}$ and any option $O \in \mathcal{O}$, we have $\mathbb{E}_\pi[S] \ge \mathbb{E}_\pi[O].$

This requires the principal to prefer delegation regardless of the decision problem: a demanding condition. Its exact requirement becomes clear through an epistemic characterization. Following \cite{dorst2021deference}, we say the principal \textbf{totally trusts} the agent if, whenever she learns the agent expects $X$ to be at least $t$, she also expects $X$ to be at least $t$. Formally:
$\mathbb{E}_\pi\!\left[ X \,\middle|\, \mathbb{E}(X) \ge t \right] \ge t$ for all random variables $X$ and thresholds $t$.

\begin{theorem}[\cite{dorst2021deference}]\label{thm:dorst}
The principal values the agent if and only if she totally trusts the agent.
\end{theorem}

This theorem reveals that universal delegation requires complete deference to the agent's expectations. Since total trust is rarely satisfied in practice, we need more nuanced conditions. The following sections develop weaker notions of delegation that apply to specific contexts and expected performance.

\subsubsection{Example}
Let $\Omega = \{a,b\}$ and $\P$ be representable as 
$$\begin{pmatrix}
    .8 & .2\\
    .1 & .9
\end{pmatrix}$$
where coordinate $i,j$ represents $P_i(\omega_j)$. Let $\pi(\{a\})=\pi(\{b\})=.5$. 

Suppose first $\mathcal{O}=\{O_1,O_2\}$, where $O_1(a)=O_2(b)=1$, and $O_1(b)=O_2(a)=-1$. Obviously $\E_\pi(O_1)=\E_\pi(O_2)=0$. But
\begin{align*}
    \E_a(O_1) &= .8-.2 =.6 & \E_b(O_1)&=.1-.9=-.8\\
    \E_a(O_2) &= -.8+.2 = -.6 & \E_b(O_2) &= -.1 + .9 = .8
\end{align*}
So, $S_a=O_1$ and $S_b=O_2$. Note that $\E_\pi(S) = \pi(\{a\})O_1(a) + \pi(\{b\})O_2(b)$. (This makes sense: if we're at world $a$, the agent will choose $O_1$, and if we're at world $b$, Bob will choose $O_2$.) So, in this decision problem, $\pi$ prefers to delegate the decision since $1>0$. 

By \cref{thm:dorst}, $\pi$ values $P$, which makes sense here because $P$'s credences are uniformly closer to the truth than $\pi$'s are at each world. 

\subsection{Uncertainty About Value}
\label{value}

In the previous section, we assumed the principal and agent were both expected utility maximizers who differed only in their beliefs. Now we allow them to have different utility functions as well, while maintaining that both are expected utility maximizers.

Following Savage (\citeyear{savage1972foundations}), we distinguish \textbf{acts} (functions $a: \Omega \to \mathcal{C}$) from \textbf{utility functions} (functions $u: \mathcal{C} \to \mathbb{R}$), where $\mathcal{C}$ is a set of consequences. An act $a$ paired with utility function $u$ induces the option $u \circ a$. When the principal has utility function $u$ and the agent has utility function $v$, they evaluate the same act $a$ through different options $u \circ a$ and $v \circ a$.

We capture this with a generalized frame:

\begin{defn}[Generalized Frame]
    A \textit{generalized frame} is a tuple $\langle \Omega, \mathcal{F}, \pi, u, \mathcal{B}, \mathbb{A}, \mathcal{C} \rangle$ where:
\begin{itemize}
\item $\langle \Omega, \mathcal{F}, \pi \rangle$ is a finite probability space (the principal's beliefs);
\item $u: \mathcal{C} \to \mathbb{R}$ is the principal's utility function  ;
\item $\mathcal{B} = \{B_\omega\}_{\omega \in \Omega}$ where each $B_\omega: 2^{\mathbb{A}} \to \mathbb{A}$ satisfies $B_\omega(\mathcal{A}) \in \mathcal{A}$;
\item and $\mathbb{A}$ is the set of all acts and $\mathcal{C}$ is the set of consequences.
\end{itemize}
\end{defn} 

Each $B_\omega$ represents a possible expected-utility-maximizing behavior of the agent. That is, there exists some utility function $V_\omega$ and probability distribution $P_\omega$ such that for any decision problem $\mathcal{A}$:
$$B_\omega(\mathcal{A}) \in \argmax_{a \in \mathcal{A}} \mathbb{E}_{P_\omega}(V_\omega \circ a)$$

The principal is uncertain which behavioral function describes the agent, reflecting her uncertainty about both the agent's utility function and beliefs. We write $(P,V)$ to refer to the random object characterizing $B$, where $P$ refers as before to the agent's actual probability function, and $V$ refers to her utilities. We write $\mathbb{E}(V(a))$ to refer to the expected value assigned to act $a$ according to the agent. 

The principal \textbf{values} the agent if, for every decision problem $\mathcal{A}$ and act $a \in \mathcal{A}$:
$$\mathbb{E}_\pi(u(B(\mathcal{A}))) \geq \mathbb{E}_\pi(u(a))$$

This says the principal prefers delegation regardless of the decision problem. When the agent shares the principal's utility function ($V_\omega = u$ for all $\omega$), this reduces to our earlier purely epistemic case. But now we can analyze delegation under both belief and value misalignment.

For a given state $\omega$, let $\left[\omega\right]\coloneqq \{\omega' \in \Omega\mid (P_\omega,V_\omega) = (P_{\omega'},V_{\omega'})\}$. That is, $[\omega]$ is the set of worlds where the agent has the same beliefs and utilities as in $\omega$. We suppose the agent has \textbf{clarity}, which ensures that if $\omega'\not\in[\omega]$, then $P_\omega(\omega')=0$. This in effect means that the agent is certain of her own beliefs and utilities. Given a few other structural assumptions about the set of acts under consideration, detailed in appendix \ref{app:proof}, we can then prove:

\begin{theorem}
\label{thm:benfav}
    Let $\langle \Omega, \mathcal{F}, \pi, u, \mathcal{B}, \mathbb{A}, \mathcal{C} \rangle$ be a generalized frame with $\mathcal{B} = (P,V)$. If $(P,V)$ is clear, then $\pi$ values $(P,V)$ if and only if 
    for any acts $a,b \in \mathbb{A}$ and every $\omega\in\Omega$, if $\mathbb{E}_\pi(u(a)\mid [\omega]) > \mathbb{E}_\pi(u(b)\mid [\omega])$ then  $\mathbb{E}_\omega(V_\omega(a)) > \mathbb{E}_\omega(V_\omega(b))$.  This further entails that for any events $X,Y$, if  $P_\omega(X)>P_\omega(Y)$, then $\pi(X\mid [\omega]) \ge \pi(Y \mid [\omega])$.
\end{theorem}

In other words, if the principal values the agent, then upon learning the agent's behavioral dispositions (what beliefs and utilities drive its choices), the principal must come to agree with the agent's preferences over acts and must be at least as confident as the agent in whatever the agent finds more probable. This is a kind of posterior alignment: the principal's conditional credences and preferences, given the agent's cognitive profile, align with what the agent itself believes and values.

Note that valuing requires the principal to prefer delegation across every decision problem under consideration. So the theorem tells us that if we can rationally delegate across the full range of decisions we're considering, then we must view the agent as having aligned beliefs and values in this conditional sense. However, if we're willing to accept more limited delegation—preferring the agent's choices only for some proper, insufficiently rich subset of decision problems—then we can potentially tolerate greater misalignment. 

For example, suppose the principal is going to face a bet on just whether it rains. She currently has credence $.5$ in Rain. She will be offered some bet---she knows not what---that returns $-\$x$ if it does not rain, and $1-x$ if it does. Suppose that $P_{\omega}(\text{Rain})=.6$, and $P_{\omega'}(\text{Rain})=.4$ and $V_\omega=V_{\omega'}=u$. Further, suppose that the principal thinks the agent is highly underconfident, so $\pi(\text{Rain}\mid [P=P_\omega])=1$, but $\pi(\text{Rain}\mid[P=P_{\omega'})=0$. It's easy to see that the principal should delegate in this situation no matter what $x$ is. She \textbf{values} the agent over the bet on just whether it will rain. But she still conditionally disagrees with the agent about what bets to take. If $x=.9$, then conditional on $[P=P_\omega]$, the principal will want to \textit{accept} the bet, while the agent will reject the bet. Thus, when we limit the scope of decisions, value can be achieved even without total agreement. 

While value provides a useful benchmark for delegation, it assumes the principal knows which decision problem she faces. In practice, delegation to AI systems occurs under uncertainty about what problems will arise. Moreover, as we'll see, an agent's capabilities can change which problems are encountered.

\subsection{Uncertainty about Reach}
\label{reach}

The value criterion developed above assumes the principal knows what decision problem she faces. But AI delegation typically happens ex ante: we must commit before specific problems arise. This introduces two complications.

First, even without value, delegation can be rational if the agent performs sufficiently well on likely problems. The principal need not prefer delegation for \textit{every} problem, only in expectation given the distribution $\mu$ over problems she might face.

Second, and more importantly, an agent's greater reach changes which problems arise. An AI system with enhanced capabilities doesn't just make better decisions---it encounters different decision problems. It has options humans lack (like sophisticated manipulation techniques) and may trigger situations humans wouldn't face (like high-stakes automated negotiations). This means the principal must compare performance across two different distributions: $\mu_{\text{self}}$ over problems she'd face, and $\mu_{\text{delegate}}$ over problems the agent would encounter.

Even if the principal values the agent she might not delegate if $\mu_{\text{delegate}}$ systematically involves worse problems than $\mu_{\text{self}}$. Conversely, a somewhat misaligned but highly capable agent might merit delegation if their reach gives access to better problems. Rational delegation thus requires comparing expected utilities across potentially different problem distributions, accounting for how capabilities shape the decision landscape itself.

The next section develops a scoring framework to formalize this comparison.

\subsection{Decision-based scoring for ex ante delegation} \label{sec:scoring}

To formalize delegation under distributions of problems, we adapt Konek's \citeyear{konek2023evaluating} scoring framework. While our framework above handles general decision problems with arbitrary sets of acts, analyzing distributions over such problems requires additional structure. For tractability, we focus on a specific but illuminating class: binary gambles.

In this restricted setting, each decision problem consists of just two acts: accepting or rejecting a gamble. Formally, a gamble $g$ is a vector of payoffs $\langle g_{\omega_1}, \ldots, g_{\omega_n} \rangle$ representing the utility the principal receives in each state if she accepts. Rejecting always yields zero. This simplification preserves the essential features of our delegation problem---the principal and agent may disagree about which gambles to accept due to different beliefs or values---while enabling us to define probability distributions over the space of possible decisions.

We can connect this to our earlier framework: each gamble $g$ corresponds to a decision problem $\{g, \mathbf{0}\}$ where $\mathbf{0}$ is the constant zero option. The principal accepts gambles with positive expected utility: $g \in D_\pi$ iff $\mathbb{E}_\pi(g) \geq 0$. When the agent has utility function $v$ and beliefs $P$, they accept based on their own assessment: $g \in D_A$ iff $\mathbb{E}_P(v \circ g) \geq 0$.

To measure the expected cost of decision errors when the specific gamble is unknown, we first need a benchmark for perfect decision-making. At any world $\omega$, the ideal set of gambles consists of those with non-negative payoff at that world:
$$\mathcal{I}_\omega = \{g \mid g_\omega \geq 0\}$$

If the principal knew $\omega$ was actual, these are exactly the gambles she should accept. The error set for any decision rule $D$ at world $\omega$ is:
$$D \Delta \mathcal{I}_\omega = (D \setminus \mathcal{I}_\omega) \cup (\mathcal{I}_\omega \setminus D)$$

This captures both Type I errors (accepting losing gambles) and Type II errors (rejecting winning gambles).

To quantify these errors, let $\mu$ be a probability measure over possible gambles the principal might face. The expected loss from using decision rule $D$ is:
$$\mathcal{L}^\mu(D) = \sum_{\omega \in \Omega} \pi(\omega) \int_{D \Delta \mathcal{I}_\omega} |g_\omega| \, \mathrm{d}\mu(g)$$

This integrates the magnitude of payoffs over the error set, weighted by both the principal's beliefs about states ($\pi$) and the distribution of gambles she expects to face ($\mu$).

\textbf{Delegation with shared distributions.} When the principal and agent face the same distribution $\mu$ of decision problems, the delegation criterion is straightforward. The principal should delegate when:
$$\mathcal{L}^\mu(D_A) \leq \mathcal{L}^\mu(D_\pi)$$

This allows delegation even with misalignment---the principal accepts that the agent will make some "wrong" decisions (from her perspective) if the overall expected loss is lower.

\textbf{Delegation with different distributions.} The important extension comes when the agent's greater reach means it faces a different distribution of problems. Let $\mu_{\text{self}}$ be the distribution the principal faces and $\mu_{\text{delegate}}$ be the distribution the agent would face.

With different distributions, we must account for both losses (errors) and gains (correct decisions). Define the gain at world $\omega$:
$$\mathcal{G}^\mu_\omega(D) = \int_{D \cap \mathcal{I}_\omega} |g_\omega| \, \mathrm{d}\mu(g)$$

The net score combines losses and gains:
$$\mathcal{S}^\mu(D) = \mathcal{L}^\mu(D) - \mathcal{G}^\mu(D)$$

The delegation criterion becomes:
$$\mathcal{S}^{\mu_{\text{delegate}}}(D_A) \leq \mathcal{S}^{\mu_{\text{self}}}(D_\pi)$$

This captures our key insight: even a somewhat misaligned but more capable agent might be worth delegating to if their expanded reach gives them access to better decision problems. Conversely, perfect alignment might not justify delegation if the agent's capabilities lead to systematically worse problems.

This framework also enables comparison across multiple agents: the principal should delegate to whichever agent achieves the best expected score given their respective distributions.

\section{Application of our Framework}

\subsection{Illustrative Example: A Noisy Expert}
\label{sec:example1}

To make our framework concrete, we first analyze a scenario where a principal, Alice, must decide whether to delegate a task to an epistemically advantaged but misaligned agent, Bob.

Alice must decide whether to \textbf{Open} a box or \textbf{Not Open} it. The box contains a payoff $\omega$ drawn uniformly from $\Omega = \{-5, 3, 8\}$. Opening the box yields $\omega$; not opening yields $0$. Bob's utility function is Alice's utility plus some unknown constant (either $+4$ or $-3$).

\begin{itemize}
    \item \textbf{Alice's Strategy ($D_\pi$):} Her expected payoff for opening is $\mathbb{E}_\pi[\omega] = \frac{1}{3}(-5+3+8) = 2$. Since $2 > 0$, her optimal strategy is to \textbf{always Open}.

    \item \textbf{Bob's Strategy ($D_B$):} Bob has an advantage and a disadvantage.
    \begin{enumerate}
        \item \textbf{Accuracy:} Before deciding, Bob gets to peek inside a \textit{different} box, revealing a value $\omega' \in \Omega$ that he knows is not in the primary box. This allows him to update his belief from the uniform prior $\pi$ to a uniform distribution over $\Omega \setminus \{\omega'\}$.
        \item \textbf{Misalignment (Noise):} Bob's payoff is different from Alice's. With 50\% probability, he is in noise state $n_1$ and for any payoff $x$ for Alice, he receives payoffs $x+4$. With 50\% probability, he is in noise state $n_2$ and receives payoffs $x-3$. His decision is based on the expected value under his updated beliefs and his own utility function.
    \end{enumerate}
\end{itemize}

We evaluate the decision to delegate using the scoring framework from \S\ref{sec:scoring}, where the net score is $\mathcal{S}(D) = \mathcal{L}(D) - \mathcal{G}(D)$ and a lower score is better. The decision space for Bob involves 12 equiprobable states, corresponding to Alice's payoff $\omega$, the peeked value $\omega'$, and the noise state $n_i$. The full, state-by-state derivation of Bob's score is provided in Appendix~\ref{app:bob_calc}.

\textbf{Alice's Score:} Since Alice always opens, she makes an error when $\omega = -5$ and is correct when $\omega \in \{3, 8\}$. Her expected scores are:
\begin{align*}
    \mathcal{L}(D_\pi) &= \pi(\omega=-5) \cdot |-5| = \frac{1}{3} \cdot 5 = \frac{20}{12} \\
    \mathcal{G}(D_\pi) &= \pi(\omega=3) \cdot |3| + \pi(\omega=8) \cdot |8| = \frac{11}{3} = \frac{44}{12} \\
    \mathcal{S}(D_\pi) &= \frac{20}{12} - \frac{44}{12} = -\frac{24}{12} = \mathbf{-2.0}
\end{align*}

\textbf{Bob's Score:} Bob's decisions depend on the information he receives and the noise state. Summing over all 12 possibilities (see Appendix~\ref{app:bob_calc}) yields his expected scores:
\begin{align*}
    \mathcal{L}(D_B) &= \frac{1}{12}(5+5+3+8) = \frac{21}{12} = 1.75 \\
    \mathcal{G}(D_B) &= \frac{1}{12}(3+3+8+8+5+5) = \frac{32}{12} \approx 2.67 \\
    \mathcal{S}(D_B) &= \frac{21}{12} - \frac{32}{12} = -\frac{11}{12} \approx \mathbf{-0.917}
\end{align*}

The criterion is to delegate if $\mathcal{S}(D_B) \leq \mathcal{S}(D_\pi)$. Since $-0.917 > -2.0$, Alice's score is better (lower), so she should \textbf{not} delegate. In this case, Bob's informational advantage does not outweigh the errors introduced by his noisy perception.

\subsection{Illustrative Example: A Misaligned Expert}
\label{sec:example2}

To explore value misalignment, we modify our scenario so Alice and Bob have different utility functions rather than epistemic noise.

Alice must decide whether to \textbf{Open} a box or \textbf{Not Open} it. The box contains a payoff drawn uniformly from $\Omega = \{-400, 25, 100, 225\}$. Opening yields the payoff; not opening yields 0. Alice maximizes monetary value linearly.

\begin{itemize}
    \item \textbf{Alice's Strategy} ($D_\pi$): Her expected payoff for opening is $\mathbb{E}_\pi[\omega] = \frac{1}{4}(-400+25+100+225) = -12.5$. Since this is negative, her optimal strategy without delegation is to \textbf{never Open}.
    \item \textbf{Bob's Strategy} ($D_B$): Bob evaluates outcomes using a risk-averse utility function: $U_B(x) = \frac{x}{\sqrt{|x|}}$. This compresses large gains while amplifying losses. Before deciding, Bob peeks at a different box to eliminate one possible value, then decides based on his expected utility. To incentivize participation, Alice pays Bob a fixed fee of \$50 when delegating.
\end{itemize}

We evaluate delegation using our scoring framework, accounting for Alice's effective payoffs after the \$50 fee. The detailed state-by-state analysis appears in Appendix B.

\textbf{Alice's Score}: Since Alice never opens, she avoids all losses but misses all gains $\mathcal{S}(D_\pi) = -\frac{150}{12}$.

\textbf{Bob's Score}: Bob's risk-averse decisions, even after Alice pays the fee, yield
$\mathcal{S}(D_B) = -\frac{700}{12}$.

Since $-700/12 < -150/12$, Bob achieves a better (lower) score, so Alice should delegate. Despite the utility misalignment and delegation fee, Bob's risk aversion shields Alice from the large negative payoff (-400) while still capturing moderate gains. This illustrates how misalignment from different risk preferences can sometimes benefit the principal when the agent's utility function aligns with avoiding the principal's worst outcomes.

\subsection{Illustrative Example: A Broader-Reach Expert}
\label{sec:reach_example}

In this section, we isolate the contribution of \textit{reach} to the delegation decision while holding both epistemic accuracy and value alignment fixed. The principal (Alice) and the agent (Bob) share identical probabilistic beliefs about the world and the same linear utility function $u(x)=x$. Both follow the same decision rule: open a box if and only if the expected payoff is non-negative, i.e., $\mathbb{E}[\omega] \geq 0$. The only difference between them lies in their \textit{reach}---the set of decision problems each can access. Bob's expanded reach allows him to act over a wider distribution of problems, potentially improving his expected score despite identical reasoning and values.

Each box represents a distinct gamble, with payoffs $\omega$ drawn uniformly from the set of possible outcomes shown in Table~\ref{tab:reach_boxes}. If Alice or Bob chooses to \textit{open} a box, they receive $\omega$; otherwise, they receive $0$. Alice's reach $\mu_{\text{self}}$ is limited to three boxes $\{A_1, A_2, A_3\}$, while Bob's reach $\mu_{\text{delegate}}$ includes two additional boxes $\{A_4, A_5\}$ unavailable to Alice. Both evaluate expected payoffs according to the same criterion.

\begin{table}[h!]
\centering
\caption{Boxes Available to Alice and Bob}
\label{tab:reach_boxes}
\renewcommand{\arraystretch}{1.2}
\begin{tabular}{ccccc}
\toprule
\textbf{Box} & \textbf{Distribution of $\omega$} & \textbf{Expected Value} & \textbf{Decision (E[$\omega$] $\geq 0$)} & \textbf{Available To} \\
\midrule
$A_1$ & $\{-6,\,3,\,9\}$ & $+2.0$ & Open & Both \\
$A_2$ & $\{-8,\,-4,\,12\}$ & $0$ & Indifferent (Open) & Both \\
$A_3$ & $\{-10,\,2,\,3\}$ & $-1.67$ & Not Open & Both \\
$A_4$ & $\{1,\,2,\,3,\,4\}$ & $+2.5$ & Open & Bob only \\
$A_5$ & $\{5,\,5,\,5\}$ & $+5.0$ & Open & Bob only \\
\bottomrule
\end{tabular}
\end{table}

We evaluate expected performance under the decision-based scoring framework introduced in Section~\ref{sec:scoring}. For any decision rule $D$ and problem distribution $\mu$, the overall score is defined as
\[
S_{\mu}(D) = L_{\mu}(D) - G_{\mu}(D),
\]
where $L_{\mu}(D)$ and $G_{\mu}(D)$ respectively represent the expected loss from incorrect decisions and the expected gain from correct ones. Delegation is rational when
\[
S_{\mu_{\text{delegate}}}(D_A) \leq S_{\mu_{\text{self}}}(D_\pi).
\]
Since Alice and Bob share the same utility and decision function, any difference in their expected performance arises solely from the change in the set of decision problems---that is, from Bob's broader reach.

Using the procedure outlined in Section~\ref{sec:scoring} and detailed in Appendix~\ref{appendix:reach_calculation}, we compute the expected loss and gain for each agent. For Alice, who has access only to boxes $A_1$--$A_3$, the aggregate loss is $L_{\mu_{\text{self}}}(D_\pi) = 2.56$ and the aggregate gain is $G_{\mu_{\text{self}}}(D_\pi) = 2.67$. Hence, her overall score is
\[
S_{\mu_{\text{self}}}(D_\pi) = 2.56 - 2.67 = -0.11.
\]
For Bob, whose reach includes boxes $A_1$--$A_5$, the aggregate loss decreases to $L_{\mu_{\text{delegate}}}(D_A) = 1.53$ and the aggregate gain increases to $G_{\mu_{\text{delegate}}}(D_A) = 3.1$, yielding
\[
S_{\mu_{\text{delegate}}}(D_A) = 1.53 - 3.1 = -1.57.
\]
Since $S_{\mu_{\text{delegate}}}(D_A) < S_{\mu_{\text{self}}}(D_\pi)$, delegation is strictly rational in expectation. This improvement in performance arises solely from Bob's expanded reach, which alters the distribution $\mu$ of problems he can face.

\subsubsection{Reach Outweighing Misalignment}

Notice that this previous example can be easily modified to show that delegation is rational, even with misalignment. Since Alice prefers to delegate to Bob by some margin, we could construct a case in which Bob charges Alice a fixed amount, less than the margin, for his delegation services. Even though he is misaligned, Alice would happily pay. 

Other cases of delegation with more much extreme version of misalignment are clearly possible as well. If Bob is likely to get a decision problem with only great payoffs for Alice, even if he tries to minimize Alice's payoffs, it can still be advantageous for Alice to delegate to him, if the best of her options is worse than the worse of his options. Of course, reach can also make things much worse, if much worse options are available to Bob. It is only one aspect of the overall deference calculus.

\section{Conclusion}

This paper provides a formal framework for AI delegation that captures the interplay between value alignment, epistemic accuracy, and reach.  Our analysis yields two key contributions. Conceptually, we identify a fundamental tension: while context-specific delegation can be rational despite misalignment, universal delegation requires near-perfect value alignment. Technically, our scoring framework extends existing approaches to handle varying problem distributions, making precise how an agent's expanded capabilities change the decision landscape, potentially making delegation attractive even with imperfect alignment if the agent accesses sufficiently better problems. 

The framework suggests that safe AI deployment requires a more subtle approach to the alignment problem. Rather than pursuing perfect alignment, we need methods to assess when a system's particular configuration of alignment, accuracy, and reach warrants delegation in specific contexts. Understanding an AI system's reach emerges as an important aspect of the delegation calculus.

Several extensions merit investigation. Most immediately, our binary gambles restriction could be relaxed to handle richer decision spaces while maintaining computational tractability. The framework could be extended to sequential decisions and non-Bayesian agents, and distributions over decision problems could depend on the state $\omega$, further endogenizing the decision context following the spirit of \cite{herrmann2023naturalizing}. For practical application, our scoring framework could formalize the empirical trade-offs measured in AI control research, providing principled methods to evaluate deployment decisions under realistic constraints.

\subsection{Online Delegation as a Learning Problem}
\label{sec:rl_section}
A particularly natural extension is to recast delegation as an online decision process. Suppose the principal repeatedly encounters environments sampled from her prior and must choose whether to act or delegate. The problem then becomes a stochastic multi-armed bandit: each arm corresponds to a choice to delegate or not, with rewards determined by realized world states.

\begin{algorithm}
\caption{Delegation Bandit $(\pi, A, T, C, u)$}\label{alg:cap}
\begin{algorithmic}[1]
\For{$t \in T$}
    \State Sample $\omega_t \sim \pi$
    \State Choose to delegate to agent $n_t=A(\{n_i,u_i\}_{i=1}^{t-1})$
    \State Receive reward $u_t=u(C_{n_t}(\omega_t))$
\EndFor
\end{algorithmic}
\end{algorithm}

In this Delegation Bandit formulation, the principal learns over time which agent yields higher expected utility, effectively updating her beliefs about alignment, accuracy, and reach through experience. Preliminary simulations using the Upper Confidence Bound (UCB) algorithm show convergence toward the optimal delegation policy across varied setups. This is unsurprising, since with finite worlds and deterministic policies, each delegation arm has a well-defined expected reward:

$$\sum_{\omega} \pi(\omega)u(D_\pi(\omega))$$

This online framing begins to connect the normative delegation calculus developed here with empirically tractable learning procedures, bridging the gap between theoretical decision models and real-world adaptive delegation. 

Of course, the extent to which we can learn our way to rational delegation through trial and error decreases as the stakes rise. While such online methods may be promising for low-stakes or repeatable contexts, the delegation calculus developed in the bulk of this paper is better suited to guiding one-shot or high-impact decisions involving sophisticated AI systems.

\begin{ack}
    We are grateful to Kevin Dorst, Jason Konek, Giacomo Molinari, and Richard Pettigrew for advice and support. We thank audiences at ILIAD and ODYSSEY for comments on earlier versions of this work. We also thank Algoverse for organizing the team of researchers, and Coefficient Giving for funding the project. B.L.\ most of his research on this project in his capacity as a faculty member at the University of Illinois. 
\end{ack}

\newpage

\bibliographystyle{chicago}
\bibliography{references.bib}

\section*{Appendices}

\appendix

\section{State-by-State Calculation for Example in \S\ref{sec:example1}}
\label{app:bob_calc}

Table \ref{tab:bob_calculation} details Bob's decision-making process and the resulting outcome for Alice across all 12 equiprobable states in the noisy expert example. The total expected Loss $\mathcal{L}(D_B)$ and Gain $\mathcal{G}(D_B)$ are derived by summing the absolute values of the outcomes, weighted by the probability of each state ($\frac{1}{12}$). An outcome is a loss if Bob opens on a negative value or fails to open on a positive value. An outcome is a gain if Bob opens on a positive value or correctly avoids opening on a negative value.

\begin{table}[h!]
\centering
\caption{Bob's Decision Process and Outcomes}
\label{tab:bob_calculation}
\renewcommand{\arraystretch}{1.2}
\begin{tabular}{ccclccr}
\toprule
\textbf{True $\omega$} & \textbf{Peeked $\omega'$} & \textbf{Noise} & \textbf{Bob's Belief} & \textbf{Noisy EU} & \textbf{Bob's Action} & \textbf{Outcome} \\
\midrule
-5 & 3  & $n_1(+4)$ & $\{-5, 8\}$ & $0.5(-1+12)=5.5$ & Open & -5 (Loss) \\
-5 & 3  & $n_2(-3)$ & $\{-5, 8\}$ & $0.5(-8+5)=-1.5$ & Not Open & +5 (Gain) \\
-5 & 8  & $n_1(+4)$ & $\{-5, 3\}$ & $0.5(-1+7)=3.0$ & Open & -5 (Loss) \\
-5 & 8  & $n_2(-3)$ & $\{-5, 3\}$ & $0.5(-8+0)=-4.0$ & Not Open & +5 (Gain) \\
\midrule
 3 & -5 & $n_1(+4)$ & $\{3, 8\}$ & $0.5(7+12)=9.5$ & Open & +3 (Gain) \\
 3 & -5 & $n_2(-3)$ & $\{3, 8\}$ & $0.5(0+5)=2.5$ & Open & +3 (Gain) \\
 3 & 8  & $n_1(+4)$ & $\{3, -5\}$ & $0.5(7-1)=3.0$ & Open & +3 (Gain) \\
 3 & 8  & $n_2(-3)$ & $\{3, -5\}$ & $0.5(0-8)=-4.0$ & Not Open & -3 (Loss) \\
\midrule
 8 & -5 & $n_1(+4)$ & $\{8, 3\}$ & $0.5(12+7)=9.5$ & Open & +8 (Gain) \\
 8 & -5 & $n_2(-3)$ & $\{8, 3\}$ & $0.5(5+0)=2.5$ & Open & +8 (Gain) \\
 8 & 3  & $n_1(+4)$ & $\{8, -5\}$ & $0.5(12-1)=5.5$ & Open & +8 (Gain) \\
 8 & 3  & $n_2(-3)$ & $\{8, -5\}$ & $0.5(5-8)=-1.5$ & Not Open & -8 (Loss) \\
\bottomrule
\end{tabular}
\end{table}

\section{State-by-State Calculation in \S\ref{sec:example2}}

This appendix details the calculation for the example in \S\ref{sec:example2} where Alice and Bob have different utility functions.

Bob's utility function $U_B(x) = x/\sqrt{|x|}$ transforms monetary payoffs into utilities that reflect risk aversion:
\begin{itemize}
\item $U_B(-400) = -400/\sqrt{400} = -400/20 = -20$
\item $U_B(25) = 25/\sqrt{25} = 25/5 = 5$
\item $U_B(100) = 100/\sqrt{100} = 100/10 = 10$
\item $U_B(225) = 225/\sqrt{225} = 225/15 = 15$
\end{itemize}

When Bob peeks at value $\omega'$ and the true box contains $\omega$, he forms a uniform belief over $\Omega \setminus \{\omega'\}$ (the three remaining values) and opens if his expected utility is positive. Alice must pay Bob \$50 to delegate, so her effective payoffs when delegating are: $-450$ (if true value is $-400$), $-25$ (if 25), $50$ (if 100), and $175$ (if 225).

Table \ref{tab:logpay} provides a compact summary of delegation outcomes. Each cell shows whether the box is a loss ($a$) or a gain ($\emptyset$), given the combination of Alice's actual payoff and Bob's perceived log payoff from peeking another box. Alice's values include the deduction of \$50 for delegation, shown in parentheses.

\begin{table}[h!]
\centering
\caption{Delegation Decisions under Logarithmic Utility with Fixed Pay (\$50)}
\label{tab:logpay}
\renewcommand{\arraystretch}{1.25}
\setlength{\tabcolsep}{6pt} 
\begin{tabular}{cccccc}
\toprule
\multicolumn{2}{c}{} & \multicolumn{4}{c}{\textbf{Bob Peeked Box}} \\
\cmidrule(lr){3-6}
\textbf{Alice's Payoff Box (with Delegation)} &
\textbf{Error Set (Alice, Bob)} &
\textbf{$-20$} & \textbf{$5$} & \textbf{$10$} & \textbf{$15$} \\
\midrule
$-400$ ($-450$) &  & $\emptyset,\,a$ & $\emptyset,\,\emptyset$ & $\emptyset,\,\emptyset$ & $\emptyset,\,\emptyset$ \\
$25$ ($-25$) & $a,\,a$ &  & $a,\,\emptyset$ & $a,\,\emptyset$ & $a,\,\emptyset$ \\
$100$ ($50$) & $a,\,\emptyset$ & $a,\,\emptyset$ &  & $a,\,a$ & $a,\,a$ \\
$225$ ($175$) & $a,\,\emptyset$ & $a,\,\emptyset$ & $a,\,\emptyset$ &  & $a,\,\emptyset$ \\
\bottomrule
\end{tabular}
\end{table}

Table \ref{tab:bob_decision_process} provides the detailed decision process for all 12 equiprobable states, showing Bob's belief formation, expected utility calculation, and the resulting outcome for Alice.

\begin{table}[h!]
\centering
\caption{Bob's Decision Process for All 12 Equiprobable States}
\label{tab:bob_decision_process}
\renewcommand{\arraystretch}{1.2}
\begin{tabular}{cccccc}
\toprule
\textbf{True $\omega$} & \textbf{Peeked $\omega'$} & \textbf{Bob's Belief} & \textbf{Bob's EU} & \textbf{Bob Opens?} & \textbf{Alice's Outcome} \\
\midrule
$-400$ & $25$  & $\{-400,\,100,\,225\}$ & $5/3$ & Yes & $-450$ (Loss) \\
$-400$ & $100$ & $\{-400,\,25,\,225\}$ & $0$ & No & $0$ (Gain) \\
$-400$ & $225$ & $\{-400,\,25,\,100\}$ & $-5/3$ & No & $0$ (Gain) \\
\midrule
$25$ & $-400$ & $\{25,\,100,\,225\}$ & $10$ & Yes & $-25$ (Loss) \\
$25$ & $100$ & $\{-400,\,25,\,225\}$ & $0$ & No & $0$ (Loss) \\
$25$ & $225$ & $\{-400,\,25,\,100\}$ & $-5/3$ & No & $0$ (Loss) \\
\midrule
$100$ & $-400$ & $\{25,\,100,\,225\}$ & $10$ & Yes & $50$ (Gain) \\
$100$ & $25$ & $\{-400,\,100,\,225\}$ & $5/3$ & Yes & $50$ (Gain) \\
$100$ & $225$ & $\{-400,\,25,\,100\}$ & $-5/3$ & No & $0$ (Loss) \\
\midrule
$225$ & $-400$ & $\{25,\,100,\,225\}$ & $10$ & Yes & $175$ (Gain) \\
$225$ & $25$ & $\{-400,\,100,\,225\}$ & $5/3$ & Yes & $175$ (Gain) \\
$225$ & $100$ & $\{-400,\,25,\,225\}$ & $0$ & No & $0$ (Loss) \\
\bottomrule
\end{tabular}
\end{table}

From Table \ref{tab:bob_decision_process}, we can calculate the expected losses and gains. A loss occurs when Bob opens on a negative value (rows 1, 4) or fails to open on a positive value (rows 5, 6, 9, 12). A gain occurs when Bob correctly opens on a positive value (rows 7, 8, 10, 11) or avoids opening on a negative value (rows 2, 3).

For Bob (with delegation), summing over all 12 equiprobable states:
\begin{align*}
\mathcal{L}(D_B) &= \frac{1}{12}(450 + 25 + 25 + 25 + 100 + 225) = \frac{850}{12} \\
\mathcal{G}(D_B) &= \frac{1}{12}(400 + 400 + 50 + 50 + 175 + 175) = \frac{1250}{12} \\
\mathcal{S}(D_B) &= \mathcal{L}(D_B) - \mathcal{G}(D_B) = \frac{850}{12} - \frac{1250}{12} = -\frac{400}{12}
\end{align*}

For Alice without delegation (never opens), she misses all positive values but avoids the negative one:
\begin{align*}
\mathcal{L}(D_\pi) &= \frac{1}{4}(25 + 100 + 225) = \frac{350}{4} = \frac{1050}{12} \\
\mathcal{G}(D_\pi) &= \frac{1}{4}(400) = \frac{100}{4} = \frac{300}{12} \\
\mathcal{S}(D_\pi) &= \mathcal{L}(D_\pi) - \mathcal{G}(D_\pi) = \frac{1050}{12} - \frac{300}{12} = \frac{750}{12}
\end{align*}

Since $\mathcal{S}(D_B) = -400/12 < 750/12 = \mathcal{S}(D_\pi)$, Bob's net score is substantially better (lower), making delegation optimal despite the \$50 fee and utility misalignment.

\section{State-by-State Calculation of \S\ref{sec:reach_example}}
\label{appendix:reach_calculation}

Both Alice and Bob follow the same decision rule: open a box if and only if the expected payoff $\mathbb{E}[\omega] \geq 0$.  
Their utilities are identical ($u(x)=x$), and beliefs about outcomes are uniform within each box.  
The only difference lies in \textit{reach}---the set of boxes each can act upon.  
Alice can choose among $\{A_1, A_2, A_3\}$, while Bob can access $\{A_1, A_2, A_3, A_4, A_5\}$.

\begin{table}[h!]
\centering
\caption{Boxes Available to Alice and Bob}
\label{tab:reach_boxes}
\renewcommand{\arraystretch}{1.2}
\begin{tabular}{ccccc}
\toprule
\textbf{Box} & \textbf{Distribution of $\omega$} & \textbf{Expected Value} & \textbf{Decision (E[$\omega$] $\geq 0$)} & \textbf{Available To} \\
\midrule
$A_1$ & $\{-6,\,3,\,9\}$ & $+2.0$ & Open & Both \\
$A_2$ & $\{-8,\,-4,\,12\}$ & $0$ & Indifferent (Open) & Both \\
$A_3$ & $\{-10,\,2,\,3\}$ & $-1.67$ & Not Open & Both \\
$A_4$ & $\{1,\,2,\,3,\,4\}$ & $+2.5$ & Open & Bob only \\
$A_5$ & $\{5,\,5,\,5\}$ & $+5.0$ & Open & Bob only \\
\bottomrule
\end{tabular}
\end{table}

For each box, Loss ($L_i$) is the average absolute magnitude of negative outcomes (if the box is opened),  
and Gain ($G_i$) is the average absolute magnitude of non-negative outcomes (if opened).  
If the box is \textit{not opened}, Loss equals the missed positive payoffs.

\begin{table}[h!]
\centering
\caption{Per-Box Loss and Gain Calculations}
\label{tab:reach_loss_gain}
\renewcommand{\arraystretch}{1.2}
\begin{tabular}{cccc}
\toprule
\textbf{Box} & \textbf{Decision} & \textbf{Loss $L_i$} & \textbf{Gain $G_i$} \\
\midrule
$A_1$ & Open & $2.00$ & $4.00$ \\
$A_2$ & Open & $4.00$ & $4.00$ \\
$A_3$ & Not Open & $1.67$ & $0.00$ \\
$A_4$ & Open & $0.00$ & $2.50$ \\
$A_5$ & Open & $0.00$ & $5.00$ \\
\bottomrule
\end{tabular}
\end{table}

Assuming uniform weight $\mu$ over each agent's reachable boxes:

\begin{align*}
L_{\mu_{\text{self}}}(D_\pi) &= \frac{2.00 + 4.00 + 1.67}{3} = 2.56, \\
G_{\mu_{\text{self}}}(D_\pi) &= \frac{4.00 + 4.00 + 0.00}{3} = 2.67, \\
S_{\mu_{\text{self}}}(D_\pi) &= L_{\mu_{\text{self}}}(D_\pi) - G_{\mu_{\text{self}}}(D_\pi) = -0.11. \\[6pt]
L_{\mu_{\text{delegate}}}(D_A) &= \frac{2.00 + 4.00 + 1.67 + 0.00 + 0.00}{5} = 1.53, \\
G_{\mu_{\text{delegate}}}(D_A) &= \frac{4.00 + 4.00 + 0.00 + 2.50 + 5.00}{5} = 3.10, \\
S_{\mu_{\text{delegate}}}(D_A) &= L_{\mu_{\text{delegate}}}(D_A) - G_{\mu_{\text{delegate}}}(D_A) = -1.57.
\end{align*}

Both agents are value- and belief-aligned; the only differentiating factor is reach.  
Bob’s expanded reach allows access to additional positively skewed gambles ($A_4$, $A_5$),  
lowering his expected loss and increasing his expected gain.  
Because $S_{\text{Bob}} < S_{\text{Alice}}$, delegation to Bob is strictly rational under identical reasoning and values.

\section{Proof of Theorem \ref{thm:benfav}}
\label{app:proof}

To state and prove the theorem, we need three further conditions.

\textbf{Richness:} If $a,a'\in\mathbb{A}$ and $X\in \F$, then $a^X_{a'}\in \mathbb{A}$, where $a^X_{a'}(\omega)=a(\omega)$ if $\omega\in X$ and $=a'(\omega)$ otherwise. 

\textbf{Stochastic Choice:} Suppose $a,b\in\A$, and  $\E_\omega(u(a))=\E_\omega(u(b))$, but $B_\omega(\A)=a$. Then there exists $\omega'$ with $\pi(\omega')>0$ such that $P_{\omega'}=P_{\omega}$ and $V_{\omega'}=V_\omega$, but $B_{\omega'}(\A)=b$.

\textbf{Constant Acts:} There are consequences $c_1$ and $c_2$ and acts $a,b$ such that (1) Alice strictly prefers $c_1$ to $c_2$, and (2) $a(\omega)=c_1$ and $b(\omega)=c_2$ for all $\omega$.

With these in hand we can now restate:

\textbf{Theorem \ref{thm:benfav}}
    Let $\langle \Omega, \mathcal{F}, \pi, u, \mathcal{B}, \mathbb{A}, \mathcal{C} \rangle$ be a generalized frame with $\mathcal{B} = (P,V)$ satisfying \textbf{stochastic choice} and \textbf{clarity}. Suppose that $\mathbb{A}$ satisfies \textbf{richness} and \textbf{constant acts}. Then $\pi$ values $(P,V)$ if and only if 
    for any acts $a,b \in \mathbb{A}$ and every $\omega\in\Omega$, if $\mathbb{E}_\pi(u(a)\mid [\omega]) > \mathbb{E}_\pi(u(b)\mid [\omega])$ then  $\mathbb{E}_\omega(V_\omega(a)) > \mathbb{E}_\omega(V_\omega(b))$.  This further entails that for any events $X,Y$, if  $P_\omega(X)>P_\omega(Y)$, then $\pi(X\mid [\omega]) \ge \pi(Y \mid [\omega])$.

\begin{proof}
    First we show that for any decision problem $\A$, $B_\omega(\A) \in \argmax_\A(\E_\pi(u(z)\mid[\omega])$. In other words, if learning we're in cell $[\omega]$ makes the principal strictly prefer $a$ to $b$, then the agent at least weakly prefers $a$ to $b$. Note that, for ease of readability, we sometimes write expressions like $u(a(\omega))$ as $u(a,\omega)$.

    Suppose not. Let $b=B_\omega(\A)$, and let $a\in \argmax_\A(\E_\pi(u(z)\mid[\omega])$. Then, 
    \[
    \sum_{\omega'\in[\omega]}P_{[\omega]}(\omega')V_{[\omega]}(b, \omega') \geq  P_{[\omega]}(\omega')V_{[\omega]}(a, \omega')
    \]
    but
      \[
      \sum_{\omega'\in[\omega]}\pi(\omega'\mid[\omega])u(a, \omega') >  \pi (\omega'\mid [\omega])u(b, \omega')
    \]
    Let $a_0$ be an arbitrary act in $\mathbb{A}$.  Define act $a^\star$ so that for any $\omega_i\in\Omega$:
    \[
        a^\star(\omega_i)\coloneqq\begin{cases}
            a_0, & \text{if } \omega_i\not\in [\omega]\\
            a, & \text{if }\omega_i\in[\omega]
        \end{cases}
    \]
    Likewise, define $b^\star$ so that:
    \[
    b^\star(\omega_i)\coloneqq
    \begin{cases}
            a_0, & \text{if } \omega_i\not\in [\omega]\\
            b, & \text{if }\omega_i\in[\omega]
        \end{cases}
    \]
    Note that $a^\star$ and $b^\star$ exist by \textbf{richness}. Let $\A' = \{a^\star, b^\star\}$. By \textbf{clarity}, $B_\omega(\A')=b^\star$, and $a^\star\in \argmax_{\A'} \E_\pi(u(z)\mid[\omega])$. Without loss of generality, assume $u(a_0,\omega_i)=0$ for all $\omega_i\in\Omega$. So,  $\E_\pi(u(B(\A'))=\E_\pi(u(b)\mid[\omega])\pi([\omega])<\E_\pi(u(a^\star))$. So, $\pi$ does not value $(P,V)$. 

    We've thus shown that if 
    \[
      \sum_{\omega'\in[\omega]}\pi(\omega'\mid[\omega])u(a, \omega') >  \pi (\omega'\mid [\omega])u(b, \omega'),
    \] we also have 

    \[\sum_{\omega'\in[\omega]}P_{[\omega]}(\omega')V_{[\omega]}(a, \omega') \geq  P_{[\omega]}(\omega')V_{[\omega]}(b, \omega').
    \]

    To make this inequality strict, we need to invoke our \textbf{stochastic choice} principle. If in $[\omega]$, the agent is indifferent between $a$ and $b$ but chooses $a$, then there is some $[\omega']$ where it has the same preferences that he does in $[\omega]$, but chooses $b$. Given $[\omega']$, we can run the same argument as above, which shows that the principal does not value the agent. 
      
    We now show that if $P_\omega(X)>P_\omega(Y)$ then $\pi(X\mid [\omega])\ge \pi(Y \mid[\omega])$. First, if $[\omega]=\{\omega\}$, then we're done. Assume $|[\omega]|>1$. Let $E\subsetneq [\omega]$, and let $F$ be another event such that $F\cap E \cap [\omega] = \emptyset$. By \textbf{constant acts} and \textbf{richness}, we can take consequences $c_1$, $c_2$ and ensure  $f(\omega')=c_1$ if $\omega'\in E$ and $c_2$ if $\omega'\in E^c$. We also let $g(\omega')=c_1$ if $\omega'\in F$ and $g(\omega')=c_2$ if $\omega'\in F^c$. If Alice prefers $c_1$ to $c_2$, then $\E_\pi(u(f)\mid [\omega]) > \E_\pi(u(g) \mid [\omega])$ if and only if $\pi(E \mid [\omega]) > \pi(F \mid \omega)$. Similarly, $\E_\omega(u(f))>\E_\omega(u(g))$ if and only if $P_\omega(E) > P_\omega(F)$. So for any events, $\pi(-\mid[\omega])$ and $P_\omega$ agree on all strict inequalities if the principal values the agent.   
    \end{proof}

    \section{Online Delegation Experiments \S\ref{sec:rl_section}}

Our experiments mainly consisted of translating the examples in Section 4 to the online delegation framework and determining whether UCB converges to the same delegation behavior expected by the static framework. We found that UCB will always converge if the variance of the box rewards are small, which is a general requirement for UCB's performance. The code we used is below.

\lstset{
        language=Python,
        basicstyle=\ttfamily\small, 
        numberstyle=\tiny\color{gray}, 
        frame=single, 
        breaklines=true, 
        captionpos=b, 
        tabsize=4, 
        showstringspaces=false 
    }

\begin{lstlisting}
import numpy as np

def sample_world(D, mu, B_behavior, num_peeked):
  # D is the set of possible boxes
  # mu encapsulates Alice's belief over what decision box she will face she's in
  # B_behavior is a function of the possible set of boxes which determines Bob's behavior
  arr = list(D)
  payoff_box = np.random.choice(arr, p=mu)
  payoff_behavior = B_behavior(D, payoff_box, num_peeked)
  return payoff_box, payoff_behavior

def B_behavior_noisy_expert(D, payoff_box, num_peeked, noise_set=set([2, -2])):
  B_D = (D - {payoff_box})
  noise = np.random.choice(list(noise_set))
  unpeeked_boxes = set(np.random.choice(list(B_D), size = len(B_D)-num_peeked, p=[1/len(B_D)]*len(B_D))+noise)
  B_D.add(payoff_box+noise)

  # print(B_D)
  e_v = 0
  for box in B_D:
    e_v += box / len(B_D)
  return True if e_v >= 0 else False

def sqrt_func(x):
  return x / np.sqrt(np.abs(x))
def B_behavior_discounted(D, payoff_box, num_peeked, discount_func=sqrt_func, payment=2):
  B_D = (D - {payoff_box})
  d_func = np.vectorize(discount_func)
  unpeeked_boxes = set(d_func(payment+np.random.choice(list(B_D), size = len(B_D)-num_peeked, p=[1/len(B_D)]*len(B_D))))
  B_D.add(discount_func(payoff_box))
  e_v = 0
  for box in B_D:
    e_v += box / len(B_D)
  return True if e_v >= 0 else False

def B_behavior_scen_2(D, payoff_box, num_peeked):
  B_D = (D - {payoff_box})
  peeked_boxes = set(np.random.choice(list(B_D), size = num_peeked, p=[1/len(B_D)]*len(B_D)))
  # print(peeked_boxes)
  B_D = B_D - peeked_boxes
  B_D.add(payoff_box)
  e_v = 0
  for box in B_D:
    e_v += box / len(B_D)
  return True if e_v >= 0 else False
  
class UCB:
    def __init__(self, n_arms):
        self.n_arms = n_arms
        self.counts = np.zeros(n_arms)  # Number of times each arm has been pulled
        self.values = np.zeros(n_arms)  # Estimated value of each arm

    def select_arm(self, t):
        # Select arm using UCB formula
        if t < self.n_arms:
            return t
        ucb_values = self.values + np.sqrt(2 * np.log(1+t*(np.log(t)**2)) / (self.counts))
        return np.argmax(ucb_values)

    def update(self, chosen_arm, reward):
        self.counts[chosen_arm] += 1
        n = self.counts[chosen_arm]
        old_value = self.values[chosen_arm]
        # Update estimated value using incremental mean formula
        self.values[chosen_arm] = ((n - 1) / n) * old_value + (1 / n) * reward

from tqdm import tqdm

# D = set([i for i in np.arange(-1, 1, 0.1)])
D = set([-5, 3, 8])
# If we don't delegate, Alice will choose the plus E.V. option based
# on her uniform belief over payoff boxes
A_behavior = sum(D) >= 0
num_peeked = 1
mu = np.ones(len(D)) / len(D)
n_arms = 2 # Either delegate or don't delegate
n_steps = 1000
correct_convergence = 0
regrets = []
arm_diffs = []
# Trial runs of UCB algorithm to determine how often UCB converges correctly
for j in tqdm(range(100)):
  choices = []
  total_reward = 0
  ucb_bandit = UCB(n_arms)
  regret = 0
  A_payoff = 0
  B_payoff = 0
  for i in range(n_steps):
    chosen_arm = ucb_bandit.select_arm(i)
    payoff_box, B_behavior = sample_world(D, mu, B_behavior_noisy_expert, num_peeked)

    if chosen_arm == 0:
      reward = payoff_box * A_behavior
      A_payoff = payoff_box*A_behavior
      B_payoff = payoff_box*(B_behavior)
    # If we delegate, then Bob may end up taking the gamble if he eliminates
    # many of the negative-payoff boxes
    else:
      reward = payoff_box if B_behavior else 0

    ucb_bandit.update(chosen_arm, reward)
    total_reward += reward
    choices.append(chosen_arm)
  if ucb_bandit.values[1] >= ucb_bandit.values[0] and B_payoff >= A_payoff:
    correct_convergence += 1
  elif ucb_bandit.values[0] >= ucb_bandit.values[1] and A_payoff >= B_payoff:
    correct_convergence += 1
  regret = sum(D) - max(A_payoff, B_payoff)
  regrets.append(regret)
  arm_diffs.append((ucb_bandit.values[1] - ucb_bandit.values[0]))

print("\nNum times UCB converged correctly: {}".format(correct_convergence))
print("\nMean regret: {}".format(np.mean(regrets)))
print("\nStd regret: {}".format(np.std(regrets)))
print("\nMean arm difference: {}".format(np.mean(arm_diffs)))
print("\nStd arm difference: {}".format(np.std(arm_diffs)))
\end{lstlisting}

\end{document}